%% file: 6090.tex
\documentclass[runningheads]{llncs}
\usepackage{graphicx}
\usepackage{comment}
\usepackage{amsmath,amssymb} % define this before the line numbering.
\usepackage{color}
\usepackage{multirow}
\usepackage{array}
\usepackage[font=small,labelfont=bf]{caption}
\usepackage[font=small,labelfont=bf]{subcaption}
\usepackage{url}
\usepackage{afterpage}

\usepackage{orcidlink}

\DeclareMathOperator*{\argmin}{arg\,min}

\begin{document}
% \linenumbers
\pagestyle{headings}
\mainmatter
\def\ECCVSubNumber{6090}  % Insert your submission number here

% \title{3D Shape Reconstruction from 2D Landmarks}
\title{A Kendall Shape Space Approach to\\3D Shape Estimation from 2D Landmarks}

\begin{comment}
\titlerunning{ECCV-20 submission ID \ECCVSubNumber} 
\authorrunning{ECCV-20 submission ID \ECCVSubNumber} 
\author{Anonymous ECCV submission}
\institute{Paper ID \ECCVSubNumber}
\end{comment}
%******************

%\begin{comment}
\titlerunning{A Kendall Shape Space Approach to 3D-from-2D Shape Estimation}
\author{Martha Paskin\inst{1}\orcidlink{0000-0003-2916-0035}\and
Daniel Baum\inst{1}\orcidlink{0000-0003-1550-7245} \and
Mason N.\ Dean\index{Dean, Mason N.}\inst{2}\orcidlink{0000-0002-5026-6216} \and \\
Christoph von Tycowicz\index{von Tycowicz, Christoph}\inst{1,3}\orcidlink{0000-0002-1447-4069}}
\authorrunning{M.~Paskin et al.}
\institute{%
Zuse Institute Berlin, Germany
\email{\{paskin,baum,vontycowicz\}@zib.de} %
\and
City University of Hong Kong, China
\email{mndean@cityu.edu.hk}%
\and
Freie Universit\"at Berlin, Germany%
}
%\end{comment}
%******************
\maketitle

\begin{abstract}
3D shapes provide substantially more information than 2D images.
However, the acquisition of 3D shapes is sometimes very difficult or even impossible in comparison with acquiring 2D images, making it necessary to derive the 3D shape from 2D images.
%
%This is particularly valuable for very large or moving objects.
%
Although this is, in general, a mathematically ill-posed problem, it might be solved by constraining the problem formulation using prior information.
Here, we present a new approach based on Kendall's shape space to reconstruct 3D shapes from single monocular 2D images.
The work is motivated by an application to study the feeding behavior of the basking shark, an endangered species whose massive size and mobility render 3D shape data nearly impossible to obtain, hampering understanding of their feeding behaviors and ecology.
2D images of these animals in feeding position, however, are readily available.
We compare our approach with state-of-the-art shape-based approaches, both on human stick models and on shark head skeletons.
Using a small set of training shapes, we show that the Kendall shape space approach is substantially more robust than previous methods and results in plausible shapes.
This is essential for the motivating application in which specimens are rare and therefore only few training shapes are available. 

\end{abstract}

\input{intro}

\input{methods}

\input{experiments}

\section{Conclusion}

In this work, we have proposed a novel, non-Euclidean approach for the estimation of 3D shapes from single monocular 2D images.
In particular, we generalized the established active shape model approach to Kendall's shape space combining concepts from geometric statistics and computational differential geometry.
The resulting geometric model provides efficient, yet highly consistent estimation of 3D shapes due to a natural encoding of constraints.

We have demonstrated the performance of our approach in experiments on the reconstruction of human poses as well as head skeletons of basking sharks.
The advantages of the proposed scheme are particularly apparent in qualitative comparison to previous approaches that rely on assumptions of linearity.
While former approaches attained lower reprojection errors, the corresponding 3D shapes show defects like non-physiological distortions indicating a need for further domain-specific regularization.
For example, the convex relaxation of the ASM proposed in~\cite{zhou20153d} developed strong  asymmetries although basis shapes are symmetric.
In contrast, our geometric method yielded highly physiological results even in challenging situations where the target shape differed greatly from the training distribution.

%* ASM approaches overfit the 2D landmarks -> regularization term needed
%* ASM convex -> asymmetry in real world shark data
%* KSS performs better for all sizes of sets of basis shapes
% * a few words about camera assumptions and how these might influence the results -> Can we say that KSS can deal better with perspective camera?

% Future work:
% * Extend the approach to more advanced camera models.
% * Tune hyper parameters
In future work, we plan to extend our approach to more advanced camera models to be able to account for strong perspective effects.
Furthermore, while we use the 2D Kendall shape space distance as reprojection error, weighted or $\ell^1$-like extensions could be employed to incorporate uncertainty in 2D landmark detection or to increase robustness against outliers, respectively.

\paragraph{Acknowledgments}
We are grateful for the funding by Deutsche Forschungsgemeinschaft (DFG) through Germany’s Excellence Strategy -- The Berlin Mathematics Research Center MATH+ (EXC-2046/1, project ID:~390685689) and by Bundesministerium f\"ur Bildung und Forschung (BMBF) through BIFOLD -- The Berlin Institute for the Foundations of Learning and Data (ref.~01IS18025A and ref.~01IS18037A).
The basking shark work was funded in part by an HFSP Program Grant (RGP0010-2020 to M.N.D.).
We also want to thank the British Museum of Natural History and Zoological Museum, University of Copenhagen for providing basking shark specimens, Pepijn Kamminga, James Maclaine, Allison Luger and Henrik Lauridsen for help acquiring CT data, and Alex Mustard (Underwater Photography) and Maura Mitchell (Manx Basking Shark Watch) for granting us permission to use their underwater images of basking sharks.

\clearpage
\bibliographystyle{splncs04}
\bibliography{6090}
\end{document}

%% file: intro.tex
\section{Introduction}

The reconstruction of a 3D shape from 2D images is an important problem of computer vision that has been the subject of research for decades ~\cite{tomasi1992shape,bregler2000recovering}.
It is a widespread challenge with sample applications including motion recognition and tracking, autonomous driving, as well as robot-assisted surgery and navigation.
%
%Generally, estimating the depth of a 3D shape from 2D images is an ill-posed mathematical problem.
Generally, estimating the 3D shape of an object from single monocular 2D images is an ill-posed problem in the sense of Hadamard.
To tackle this problem, several approaches have been proposed, the choice of which greatly depends on the type and amount of data, and the application.

One common approach is to use a series of 2D images~\cite{howe2004silhouette,cao20133d} or multiple cameras~\cite{larsen2007temporally,jiang20123d,mustafa2015general}.
This is used, for example, in photogrammetry.
In order to calculate the depth, corresponding points in 2D images need to be identified~\cite{plankers2001tracking,cao20133d,agudo2014good}.

Another approach is based on physical modeling of the object to be estimated~\cite{sanzari2016bayesian,akhter2015pose}.
Here, the object is described as a kinematic tree consisting of segments connected by joints.
Each joint contains some degrees of freedom (DOF), indicating the directions it is allowed to move. 
The object is modeled by the DOF of all joints. 
Length constraints can also be added, such as limb lengths constraints in a human pose.
This helps reduce the depth ambiguity by constraining the relative position of the joints.

In lieu of additional viewpoints or physical models, another line of work regularizes the estimation problem by employing prior knowledge learned from examples of the respective object class.
Some works in this category are based on the concept of shape spaces.
Here, the 3D shape of an object is estimated by interpolating through a set of pre-defined 3D shapes, called training shapes, that represent possible deformations of the object. 
Given a 2D projection of the object, the aim is to find the ideal camera parameters and the shape coefficients describing the interpolation.
Most commonly, \emph{active shape model} (ASM) techniques are used~\cite{mori2006recovering,hejrati2012,ramakrishna2012,wang2014,zia2013,zhou20153d}.
ASM has been used intensively for the purpose of recognizing and locating non-rigid objects in the presence of noise and occlusion~\cite{heimann2009SSMreview,ambellan2019statistical}.
It aims at providing a robust approach, which accommodates shape variability by arguing that when the objects deform in ways characteristic to the class of objects they represent, the method should be able to recognize them.
Here, the \emph{shape} of a class of objects is described by a set of labeled \emph{landmark} points, each representing a particular part of the object or its boundary.
Given a set of pre-defined \emph{basis shapes} of an object, the assumption is that any shape of the object can be described by linearly interpolating between basis shapes.
The legitimacy of the model is determined by the set of training shapes, which needs to include the variability in the shape of the object.

Apart from these more classical approaches, recent years have mostly been devoted to the development of deep learning-based methods that exploit domain-specific data repositories with abundant samples (e.g., \textit{ShapeNet}, \textit{Human3.6M}).
In 2017, a kinematic skeleton fitting method using a \emph{convolutional neural network} (CNN) was proposed that regresses 2D and 3D joint positions jointly in real-time~\cite{mehta2017vnect}.
More recently, \emph{Procrustean regression networks} (PRN) were proposed~\cite{park2020procrustean}.
Using a novel loss function, they solve the problem of automatically determining the rotation.
Another recent deep learning-based method, \emph{KShapeNet}~\cite{friji2020kshapenet}, makes use of Kendall's shape space by modeling shape sequences first as trajectories on Kendall's non-linear shape space, which are then mapped to the linear tangent space.
The authors report~\cite{friji2020kshapenet} that the non-linear motion can be captured better using their approach than using previous geometric deep-learning methods.

In general, the training of deep neural network models requires a considerable amount of data.
While the generalization of learned models to new classes with few available examples has become a highly popular research topic (see e.g.~\cite{michalkiewicz2020few}), such base models are rarely available in many applications.
There is therefore a strong impetus to develop so-called shallow learning approaches that can be applied even if few-shot learning is not an option.

Our work contributes to this area by deriving a novel shape space-based approach.
We demonstrate the effectiveness of the proposed scheme in an application to the study of marine animal behavior.
In this application, we are faced with the retrieval of the 3D shape of a non-rigid (biological) object — the head skeleton of a basking shark — from single monocular 2D images.
Basking sharks (\textit{Cetorhinus maximus}) are massive animals and thought to be one of the most efficient filter-feeding fish in terms of the throughput of water filtered through their gill rakers~\cite{sanderson2016fish}.
Details about the underlying morphology of the basking shark filter apparatus and particularly how it is configured when the animal feeds have not been studied due to various challenges in acquiring real-world data.
In this study, we address this issue, using several computed-tomography (CT) scans to create skeletal representations of basking shark head structure.
From these, and known aspects of the biology (e.g.\ shark cranial anatomy, feeding  behaviors), we semi-automatically created plausible configurations of the skeleton which we use as training shapes to describe the shape space.
The ultimate goal is, given a photograph of a basking shark in feeding position, to estimate a plausible 3D shape of the skeletal representation from the training data and annotations of the skeletal joints in the form of 2D landmarks.
\begin{figure}[t]
    \centering
    \includegraphics[width=\textwidth]{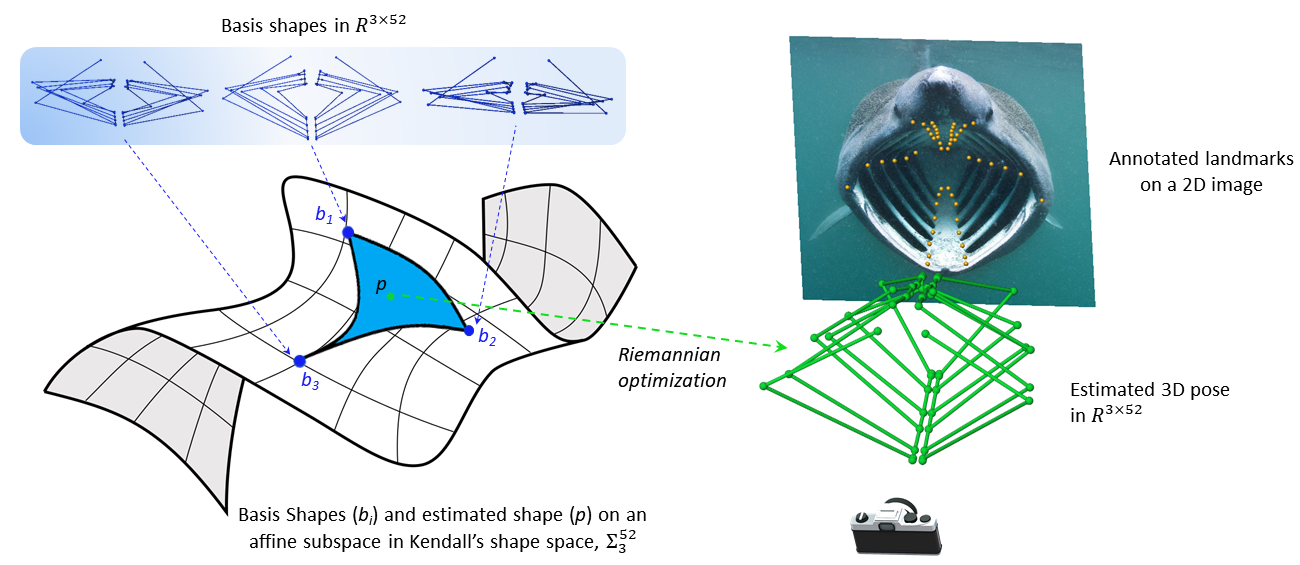}
    \caption{Methodology diagram for the proposed method.}
    \label{fig:methodology}
\end{figure}

Since the number of training shapes is very limited in our application, deep learning-based methods cannot be applied.
When using only a single 2D image, photogrammetry-based methods also cannot be applied.
And for a physical modeling-based approach, information about the movement of the skeletal elements that is supported by the joints would be necessary.
This information is not available.
We therefore resort to shape space-based methods.
In particular, we develop a new \emph{Kendall shape space}~(KSS) approach for the reconstruction of 3D shape from 2D landmarks.
Our overall workflow is sketched in Fig.~\ref{fig:methodology}.
In order to evaluate our approach, we compare it qualitatively and quantitatively with state-of-the-art ASM approaches~\cite{zhou20153d} on both human poses and basking shark skeletons.
Due to the limited number of training shapes for the basking sharks, we are mainly interested in the performance of the approaches w.r.t.\ a small number of training shapes.

The results of our experiments suggest that our method is substantially more robust than previous shape space-based methods.
This is particularly true for real world data both of human beings and shark data.
It is also computationally very efficient and provides plausible interpolations without auxiliary regularization since the estimated shape will always lie on a subspace in Kendall's shape space spanned by the training data.

The source code of the proposed method together with the derived basking shark shapes have been made publicly available~\cite{Code}.

%TODO: Add some short description of the results.
%* Low error achieved compared to ASM approaches in case of noiseless data.
%* Biological application provided
%* Better qualitative performance for both human and shark data
%* Fastest run-time (with the hyper-parameters used in this paper)

%% file: methods.tex
\section{Methods}\label{sec:methods}

In this work, we approach the problem of estimating the shape of 3D objects from single images.
To this end, we employ a landmark-based representation, viz.\ a set of ordered 3D points $(x_i)_{i=1,\ldots,k}$ that correspond to homologous points (e.g.\ joint centers in a skeleton) common to all subjects belonging to the object class under study.
By assuming the weak perspective camera model, the projection or forward map is given by the linear system
\begin{equation} \label{eq:projection}
    W = \Pi R X + t\mathbf{1}_k^T,
\end{equation}
where $X\in\mathbb{R}^{3 \times k}$ and $W\in\mathbb{R}^{2 \times k}$ denote the matrices of stacked landmarks and their 2D projections, respectively.
The rotation $R\in SO(3)$ and translation $t\in\mathbb{R}^2$ encode the camera coordinate system while the projection is carried out by the matrix
\[
\Pi = \left(
\begin{array}{ccc}
    \alpha & 0 & 0 \\
     0 & \alpha & 0
\end{array}
\right),
\]
where the scalar $\alpha$ depends only on the focal length and distance to the object.
The inverse problem of finding the 3D configuration corresponding to given projections is commonly formulated in terms of a least-squares problem.
However, the underdeterminedness renders such a problem ill-posed requiring inclusion of prior knowledge in order to obtain feasible solutions.
An established strategy based on the active shape model~\cite{cootes1995active} is to restrict the search space to linear combinations of basis shapes $B_1,\ldots,B_n \in \mathbb{R}^{3 \times k}$, i.e. 
\begin{equation}
    \label{eq:linear_blend}
    \textstyle
    X = \sum_{j=1}^n c_j B_j,
\end{equation}
%where $c_j\in\mathbb{R}$ denotes the weight of the $j$-th basis shape.
with weights $c_j$ satisfying the constraint $\sum_j c_j = 1$.
The quality of this approach, however, depends on the validity of the assumption that instances of the object class are well approximated by a hyper-planar manifold.
In general, the shape of an object is considered to be invariant under rotation, translation and scaling and, hence, takes values in a nonlinear space with quotient structure.
In particular, for landmark-based representations, this concept gives rise to the well-known Kendall shape space~\cite{kendall1984shape}.
The non-Euclidean nature of shape space not only violates the linearity assumption but furthermore prevents direct application of algebraic expressions such as Eq.~(\ref{eq:linear_blend}) due to the lack of a global vector space structure.
Nevertheless, the framework of (computational) differential geometry provides us with a rich toolset to generalize the ASM approach for use within the reconstruction problem.

\subsection{Geometric model}

Before we formulate a manifold version of the least-squares estimation, we first recall essential notions and properties from Kendall's shape space.
We refer the reader to the comprehensive work~\cite{kendall2009shape} for an in-depth introduction.

Let $X\in\mathbb{R}^{m\times k}$ be the matrix holding $k$ landmarks $x_i \in \mathbb{R}^{m}$ as columns.
We remove the effect of translations and scaling by subtracting the row-wise mean from each landmark and then dividing $X$ by its \emph{Frobenius norm} (denoted $\|\cdot\|$).
The set of all such configurations constitutes the \emph{pre-shape space}
\begin{equation*}
    \textstyle
    \mathcal{S}^k_m=\{X \in \mathbb{R}^{m \times k} \: : \sum_{i=1}^k x_i=0 \: , \, \| x\|=1 \}.
\end{equation*}
Now, the left action of $SO(m)$ on $\mathcal{S}^k_m$ given by $(R, X) \mapsto RX$ defines an equivalence relation given by $X \sim Y$ if and only if $Y = RX$ for some $R \in SO(m)$.
Kendall's shape space is defined as the quotient $\Sigma^k_m = \mathcal{S}^k_m /\mathord{\sim}$.
Now, denoting the canonical projection of $\sim$ by $\pi$ and the spherical distance by $d_\mathcal{S}(X,Y)= \text{arccos}\left< X,Y \right>$, the induced distance between any two shapes $\pi(X)$ and $\pi(Y)$ is given by
\begin{equation} \label{eq:dist}
    \textstyle
    d_{\Sigma}(X,Y):= \underset{R\in SO(m)}{\min} d_\mathcal{S}(X,RY) = \text{arccos}\sum_{i=1}^n \lambda_i,
\end{equation}
where $\lambda_1 \geq \lambda_2 \geq \dots \geq \lambda_{m-1} \geq |\lambda_m|$ are the pseudo-singular values of $YX^T$~\cite{nava2020geodesic}.
The points $X$ and $Y$ are said to be \emph{well-positioned} (denoted by $X \overset{w}{\sim} Y$), if and only if, $YX^T$ is symmetric and $d_\mathcal{S}(X,Y)=d_{\Sigma}(X,Y)$.
Note that for $m\geq3$, the shape space $\Sigma^k_m$ contains singularities making it a strata of manifolds with varying dimensions.

%In order to derive a geometric version of Eq.~(\ref{eq:linear_blend}), we first note that the 

In order to generalize the ASM approach, we first note that Eq.~(\ref{eq:linear_blend}) effectively parametrizes the affine subspace spanned by the basis shapes.
A common approach to obtain equivalent notions in Riemannian manifolds is to define \emph{geodesic subspaces} that are spanned by generalized straight lines (geodesics) emanating from a reference point and with directions restricted to a linear subspace of the tangent space~\cite{fletcher2004principal}.
However, as geodesic subspaces rely on tangent vectors, they can only be defined within regular strata of the shape space.
Therefore, we opt for an alternative construction by Pennec~\cite{pennec2018barycentric} called \emph{barycentric subspaces}, which are defined as the locus of points that are weighted means of the basis shapes.
Specifically, let $b_1 = \pi(B_1),\ldots,b_n = \pi(B_n)\in\Sigma^k_m$ be a set of $n$ distinct basis shapes, then the \emph{Fr\'echet barycentric subspace} is given by
\begin{equation*}
    FBS_{(b_j)_j} = \left\{ \argmin_{p\in\Sigma^k_m} \sum_{j=1}^n c_j \, d^2_\Sigma(p,b_j) : \sum_{j=1}^n c_j \neq 0 \right\}.
\end{equation*}

With a notion of affine submanifold at hand, we next derive a reprojection error that is defined on Kendall's shape space. 
Consistently to the 3D case, we will treat the 2D projections as points in shape space $\Sigma^k_2$ and employ the geodesic distance as objective for the reconstruction problem.
To this end, we re-formulate the projection (\ref{eq:projection}) as a map from 3D KSS to 2D KSS by
\begin{equation} \label{eq:projection_Kendall}
    p \in \Sigma^k_3 \mapsto \pi\circ\nu(R_{[1,2]}\pi^\dagger(p)) \in \Sigma^k_2,
\end{equation}
where $\nu$ performs normalization (i.e.\ projects onto the sphere) and the subscript~$[1,2]$ refers to the first two rows of $R \in SO(3)$.
To assure class invariance under $\sim$, we introduce the map $\pi^\dagger$, which yields a unique representative for a shape by well-positioning to a fixed reference pre-shape $X \in \mathcal{S}^k_3$, viz.
\begin{equation}
    \pi^\dagger : p \in \Sigma^k_m \mapsto P \in \mathcal{S}^k_m ~~ s.t. ~ \pi(P) = p, P \overset{w}{\sim} X.
\end{equation}
Note that due to the quotient structure, the re-formulation no longer depends on the scaling and translation camera parameters $\alpha$ and $t$ found in the original formulation (\ref{eq:projection}).

We are now in the position to obtain a geometry-aware counterpart of the estimation problem in terms of the projection (\ref{eq:projection_Kendall}) and the Fr\'echet barycentric subspace resulting in the following optimization problem
\begin{equation} \label{eq:3Dfrom2D}
    \min_{p, R} d_\Sigma^2\left(W,\nu(R_{[1,2]}\pi^\dagger(p))\right) ~~ s.t. ~ p \in FBS_{(b_j)_j}, R \in SO(3).
\end{equation}

\subsection{Algorithmic treatment}

In order to solve the reconstruction problem (\ref{eq:3Dfrom2D}), we employ an alternating optimization strategy taking advantage of the product structure of the search space, viz.\ $FBS_{(B_j)_j} \times SO(3)$.
Each outer iteration thus consists of two steps: First, rotation $R$ is minimized while considering the current shape estimate $p$ fixed.
To this end, we perform Riemannian optimization on the \textit{Stiefel} manifold $$V_2(\mathbb{R}^3) := \{X \in \mathbb{R}^{3 \times 2} : X^TX = I_2\} \cong SO(3)$$ utilizing the \texttt{Pymanopt}~\cite{boumal2014manopt} library.
Second, we fix the rotation $R$ and solve the resulting sub-problem w.r.t.\ shape $p$ parametrized by barycentric weights.
For Riemannian computations on Kendall's space space, we employ \texttt{Morphomatics}~\cite{Morphomatics}.
As both steps are guaranteed to weakly decrease the reconstruction objective, we obtain a simple, yet convergent optimization scheme.
In particular, both sub-problems are solved employing steepest descent strategies, where gradients are computed via automatic differentiation with \texttt{JAX}~\cite{frostig2018compiling}.
% TODO: initial guess

Due to the non-Euclidean structure of the shape space, there are no closed-form expressions for determining the (weighted) Fr\'echet mean of a given collection of sample points.
Here, we adopt an efficient, inductive estimator to derive a parameterization of the Fr\'echet barycentric subspace that exhibits only a linear computational cost in the number of basis shapes.
In particular, let $b_1,\ldots,b_n \in \Sigma^k_3$ and $w_1,\ldots,w_n \in \mathbb{R}$ such that $\sum_{j=1}^n w_j \neq 0$, then the parameterization $(w_j)_j \mapsto \mu_n \in \Sigma^k_3$ is given by the recursion
\begin{equation}
    \mu_1 = b_1 ~~~~~~~~~~ \mu_j = \gamma_{\mu_{j-1}}^{b_j}\left(\frac{w_j}{\sum_{l=1}^j w_l}\right),
\end{equation}
where $\gamma_{\mu_{j-1}}^{b_j}$ denotes the shortest geodesic from $\mu_{j-1}$ to $b_j$.
Note that under certain conditions (cf.~\cite{chakraborty2020manifoldnet}) the recursion can be shown to converge to the actual weighted Fr\'echet mean.
In any case, we expect the Fr\'echet barycentric subspace to be contained in the image of the above parameterization.

%% file: experiments.tex
\section{Experiments}
\label{sec:experiments}

We perform qualitative and quantitative evaluation of the proposed Kendall shape space (\emph{KSS}) approach in experiments concerning the estimation of human poses as well as head skeletons of basking sharks.
Throughout the experiments, we additionally provide results for the vanilla ASM approach as well as a recent convex relaxation thereof presented by Zhou et al.~\cite{zhou20153d}.
We will refer to the former and latter as \textit{ASM non-convex} and \textit{ASM convex}, respectively.

\subsection{Human Pose Estimation}

% In order to be directly able to compare our results to previous ones, we use a setup inspired by the publication by Zhou et al.~\cite{zhou20153d}, which uses poses from a MoCap dataset~\footnote{Carnegie Mellon University - CMU Graphics Lab - motion capture library http://mocap.cs.cmu.edu/} for defining training and test shapes.

The design of this experiment closely follows the setup proposed in~\cite{zhou20153d} in order to provide comparability.
In particular, we employ the MoCap dataset\footnote{Carnegie Mellon University - Graphics Lab - motion capture library \url{http://mocap.cs.cmu.edu/}} for defining training and test shapes.
Sequences from subject 86 are used as training data and sequences from subjects 13, 14 and 15 are used as testing data.
All sequences are composed of a variety of poses, including running, climbing, jumping, and dancing, and can hence be used to define a comprehensive set of basis shapes.
While 41 landmarks are used to encode human poses in the MoCap database, we use a reduced representation comprising a subset of 15 landmarks mimicking the construction in~\cite{zhou20153d}.
To reduce the effects of rotation and translation as much as possible for the Euclidean approaches, we perform generalized Procrustes analysis of the data during pre-processing.

As this dataset features abundant training data, we apply a strategy to learn an appropriate set of a restricted number of basis shapes.
An immediate choice to determine such a set would be to perform barycentric subspace analysis~\cite{pennec2018barycentric}, thus finding an optimal subspace w.r.t.\ unexplained variance.
However, as this could introduce a positive bias for our approach when comparing to the Euclidean ones, we opt for the established k-means clustering~\cite{lloyd1982least}.
Specifically, we set the basis shapes as the arithmetic means of each determined cluster thereby providing a homogeneous sampling according to the Euclidean metric.

% To choose a set of basis shapes from the thousands of available shapes, we employ k-means clustering~\cite{lloyd1982least}.
% %
% The cluster centroids offer a good representation of each cluster and are therefore chosen as basis shapes.
% %
% We are mainly interested in how the different approaches perform using a comparably small number of basis shapes.
% %
% Yet, we are interested how the performance of the approaches depends on the number of basis shapes used.
% %
% Therefore, we create sets of 32, 64 and 128 basis shapes, respectively.

\subsubsection{Quantitative Testing}

\begin{figure}[t]
    \centering
    \includegraphics[width=\textwidth]{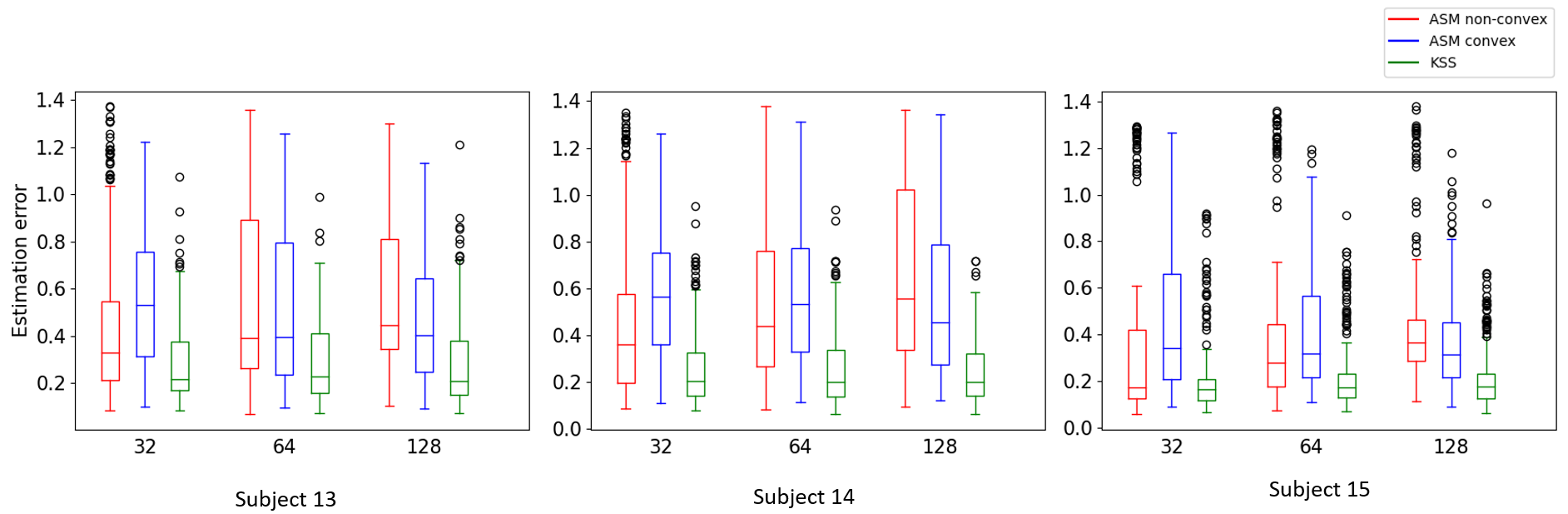}
    \caption{Boxplots of shape estimation errors of human pose test shapes using the three approaches.}
    \label{fig:human_testing}
\end{figure}
\begin{table}[b]
    \parbox{\linewidth}{
    \centering
    \caption{Mean(variance) estimation errors for human pose test shapes.}
    \label{tab:human_errors}
    \begin{tabular}{|c|c|c|c|c|}
        \hline
         &\#Basis shapes & 32 & 64 & 128 \\ 
         \hline
         \rule{0pt}{0.25cm} \multirow{3}{*}{Subject 13}& ASM NC & 0.462(0.121) &0.548(0.142) & 0.571(0.104) \\
         & ASM Conv & 0.550(0.071) &0.513(0.100) & 0.461(0.059) \\
        & KSS & \textbf{0.295(0.031)} & \textbf{0.295(0.031)} &  \textbf{0.288(0.038)} \\
         \hline
         \rule{0pt}{0.25cm}\multirow{3}{*}{Subject 14} &ASM NC &0.475(0.129) &0.544(0.127) & 0.644(0.128) \\
         & ASM Conv & 0.583(0.066) &0.566(0.077) & 0.533(0.087) \\
         & KSS & \textbf{0.267(0.029)} &  \textbf{0.258(0.028)} &  \textbf{0.242(0.019)} \\
         \hline
         \rule{0pt}{0.25cm}\multirow{3}{*}{Subject 15} & ASM NC &0.358(0.136) &0.406(0.122) & 0.457(0.089) \\
         & ASM Conv &0.450(0.092) & 0.416(0.069) & 0.366(0.043) \\
         & KSS & \textbf{0.221(0.034)}  & \textbf{0.231(0.027)} & \textbf{0.221(0.021)}  \\
         \hline
    \end{tabular}
    }
\end{table}
Similar to the creation of basis shapes, 200 shapes were chosen for each of the test subjects by performing k-means clustering on the pose sequences.
The shapes were then projected along the $z$-direction to create the 2D landmarks.
We then estimated the 3D shapes from the 2D landmarks for all approaches with different numbers of basis shapes (32, 64, and 128) and computed for each reconstruction the estimation error as the Procrustes distance~(\ref{eq:dist}) between the true and estimated shapes.
Fig.~\ref{fig:human_testing} shows boxplots of estimation errors obtained for the derived test shapes for the ASM non-convex, ASM convex, and our approach (listed as KSS), respectively.
Table~\ref{tab:human_errors} gives an overview of the error distribution. Statistical Friedman tests were conducted for the different combinations of test subjects and number of basis shapes to examine the effect that the reconstruction methods had on estimation errors. Results showed that the methods lead to statistically significant differences in estimation errors ($p < 0.001$).
The Kendall shape space approach out-performs the other approaches by estimating the unknown 3D shapes with consistently lower errors for all sets of basis shapes and test subjects.
The accuracy of the approach is directly proportional to the number of basis shapes used.

\subsubsection{Qualitative Testing}

\begin{figure}[t]
    \centering
    \includegraphics[width=9cm]{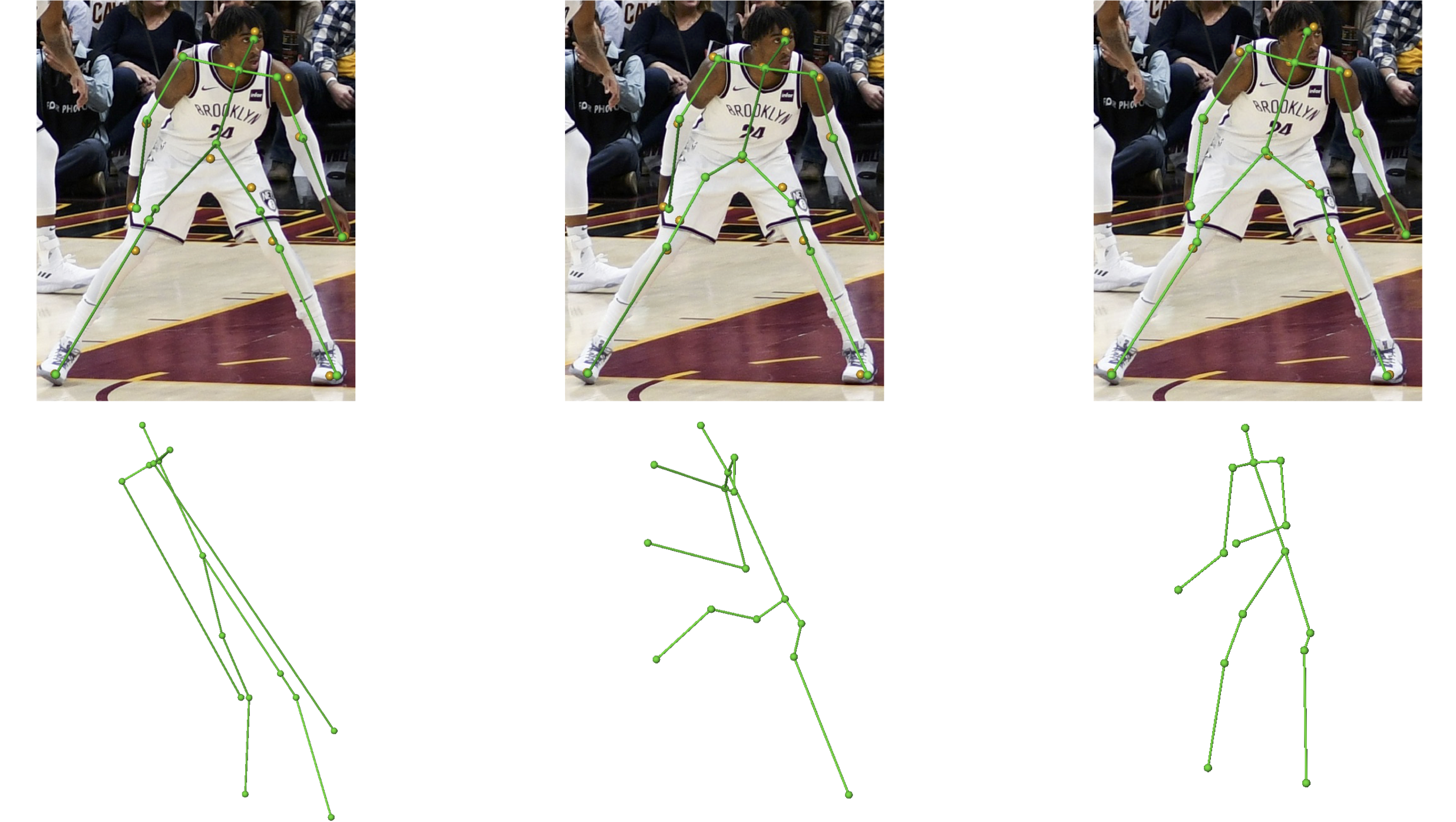}
    \caption{Comparison of the three approaches for 3D shape estimation applied to an image of a basket ball player. Top row: Fitting of the estimated shapes (green) to the landmarks (yellow). Bottom row: side views of the estimated shapes. Left: ASM non-convex approach. Middle: ASM convex approach. Right: KSS approach (ours).}
    \label{fig:man}
\end{figure}

Next, we investigate the performance of the approaches in estimating the 3D pose of people in real-world images with manually labelled landmarks.
Fig.~\ref{fig:man} shows the estimated 3D poses of a basket ball player from different viewpoints using all three approaches.
The ASM approaches perform very well in estimating a 3D shape that fits to the 2D landmarks, but the estimated shapes are implausible, which can be seen by viewing them from the side.
Moreover, they are vastly different from the basis shapes.
The proposed approach, although compromising an exact fit to the 2D landmarks, is able to estimate the 3D pose with greater plausibility.
This may be attributed to more consistent interpolation between the basis shapes adhering to the Kendall shape space constraints, %and lower degrees of freedom
which lead to reduced distortion of the basis shapes.

\begin{figure}[t]
    \centering
    \includegraphics[width=9cm]{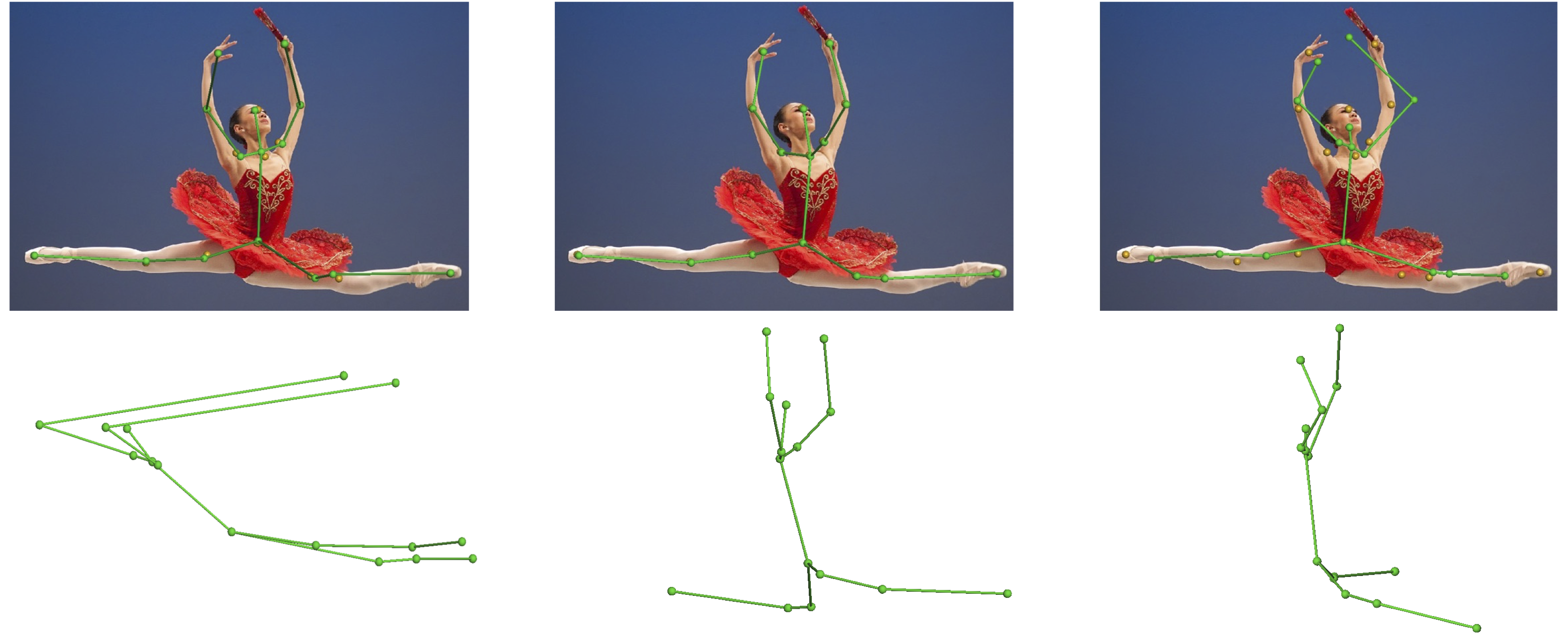}
    \caption{Comparison of the three approaches for 3D shape estimation applied to an image of a ballet dancer. Top row: Fitting of the estimated shapes (green) to the landmarks (yellow). Bottom row: side views of the estimated shapes. Left: ASM non-convex approach. Middle: ASM convex approach. Right: KSS approach (ours).}
    \label{fig:ballet}
\end{figure}

A similar scenario is shown in Fig.~\ref{fig:ballet}.
Here, however, even though the KSS results in a plausible shape, it does not fit to the shape we would expect to obtain.
This is most likely attributed to the fact that the ballet pose cannot be estimated from the given basis shapes.

\subsection{Estimating Head Skeletons of Basking Sharks}
\begin{figure}[t]
    \centering
    \begin{subfigure}{\textwidth}
    \centering
    \includegraphics[width=8cm]{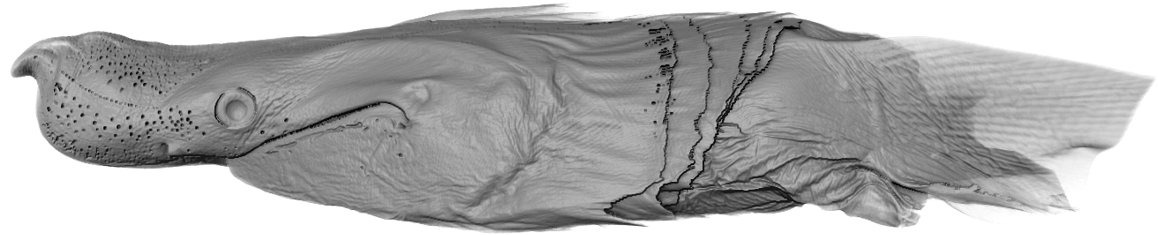}
    \caption{CT scan of a basking shark (BS3) in left lateral view.}
    \end{subfigure}
    \begin{subfigure}{\textwidth}
    \centering
    \includegraphics[width=\textwidth]{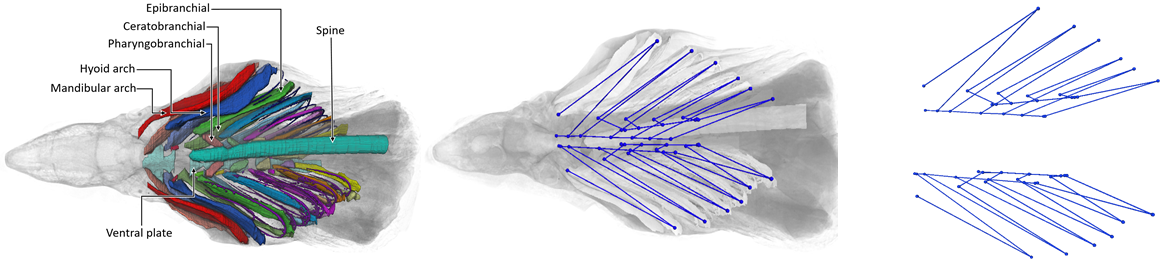}
    \caption{(left)~Segmented CT scan of the head skeleton of a basking shark (BS3, ventral view). (middle)~Piece-wise linear skeleton created by selecting end points of each region.(right)~Skeleton split into right and left halves.}
    \end{subfigure}
    \begin{subfigure}{\textwidth}
    \centering
    \includegraphics[width=8cm]{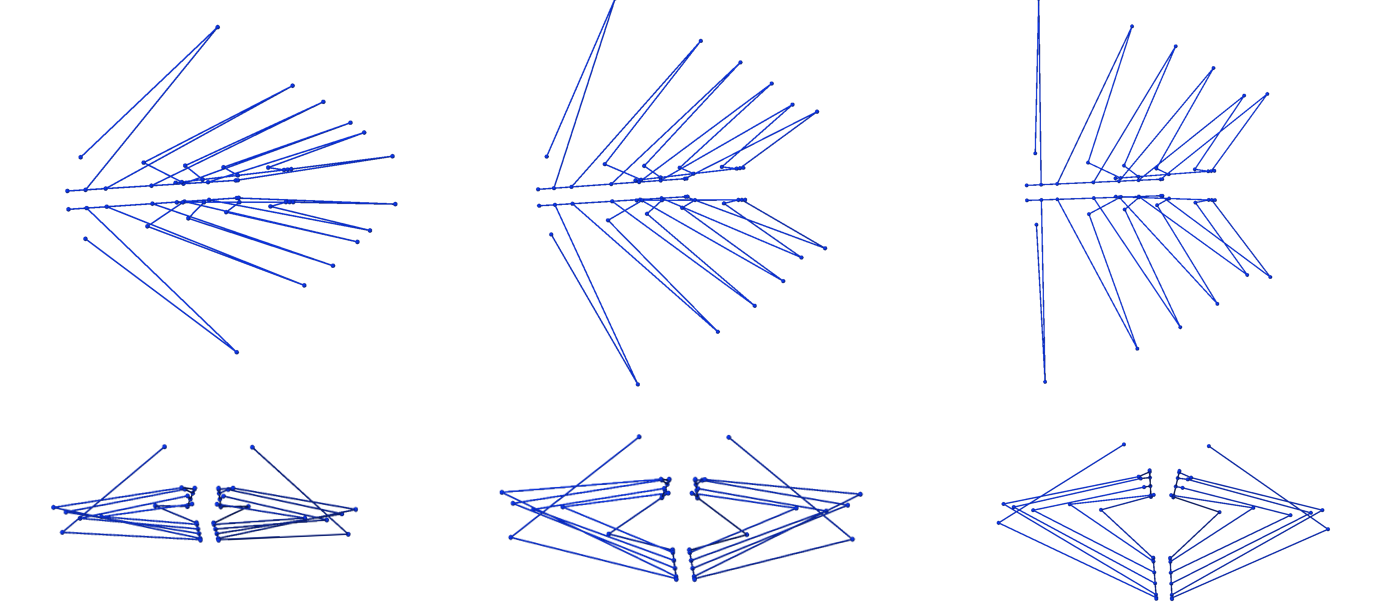}
    \caption{Basking shark skeleton shapes created by mirroring optimized half skeletons. (top)~Top/dorsal-view showing the expansion of gill arches. (bottom)~Front/anterior-view showing varying degrees of openness of the mouth.}
    \end{subfigure}
    \caption{From CT scan to skeletal representation of shark head. (a)~Volume rendering of CT scan. (b)~Segmentation, skeletonization, and separation of left and right parts of the skeleton. (c)~Opening of skeleton to create open-mouth configurations of the head.}
    \label{fig:bs_skeleton_creation}
\end{figure}

In this section, we give the results for the application that motivated the new approach for shape estimation, that is, the estimation of the shape of the head skeletons of basking sharks in feeding position.
As specimens are hard to come by and the species is heavily protected under conservation laws in many countries, the ability to extract the 3D structure of the head region non-invasively, using real-world underwater images is a great contribution to the field of biology.
We use computed tomography~(CT) scans of the heads of three basking sharks (\textit{Cetorhinus maximus}) from two museum collections (BMNH: British Museum of Natural History; ZMUC: Zoological Museum, University of Copenhagen); sample and scan information are as follows: BS1: BMNH 1978.6.22.1, size: $512 \times 512 \times 1389$;
BS2: BMNH 2004.4.15.30, size: $512 \times 512 \times 1654$;
BS3: ZMUC (no accession number), size: $512 \times 512 \times 1862$.
BS1 and BS2 were scanned at the Royal Brompton Hospital, BS3 at the Aarhus University Hospital.
CT data were used to derive basis shapes and test the methods on real-world images of the species with manually labelled landmarks.
BS3 is shown in Fig.~\ref{fig:bs_skeleton_creation}a.

Basking sharks feed by holding their mouths agape and ballooning the pharynx (throat region), allowing water to flow out the gill slits on the sides of the head~\cite{sanderson2016fish,sims2008sieving,wegner2015elasmobranch}.
The pharynx is supported by jointed branchial arches, comprising rod-like skeletal elements linked together to form a mobile skeletal basket around the throat, which can expand and collapse.
The kinematics of the pharyngeal skeleton are key to the filtering mechanism, but cannot be visualized from outside the body.
From the CT scans of basking sharks, the relevant elements of the head skeleton (the jaw, hyoid and branchial arches, skull, and spine) are isolated using manual and automatic segmentation techniques.
These isolated elements are then used to create a piece-wise linear skeleton of the sharks' pharyngeal region by labelling the start and end points of each skeletal element (e.g.\ the rod-like ceratobranchial cartilages; Fig.~\ref{fig:bs_skeleton_creation}b) and connecting each point pair via a linear segment.
Due to preservation conditions (e.g.\ specimens fixed in non-physiological positions), sharks in the CT scans showed non-physiological deformations (e.g.\ the branchial basket collapsed asymmetrically under the shark's weight), structures needed correction before creating the set of basis shapes.
Due to the bilateral symmetry of basking sharks, their right and left halves are essentially identical: skeletons were split into right and left halves which could later be mirrored to create full skeletons.
A set of optimization problems which aim to conserve the lengths of segments were used to move the spine and the ventral plate (a portion of the branchial basket) of the half skeletons into an anatomically plausible position~\cite{wegner2015elasmobranch}.
The mouth of the skeleton was opened using the same method.
The corrected skeletons were then mirrored w.r.t.\ the sagittal plane.
Fig.~\ref{fig:bs_skeleton_creation}c visualizes this process.
The anatomical arrangements of the final skeletons were approved by experts in the field (M.~Dean \& F.~Mollen, personal communication).

\subsubsection{Analysis of reconstruction accuracy}

\begin{figure}[t]
    \centering
    \includegraphics[width=9cm]{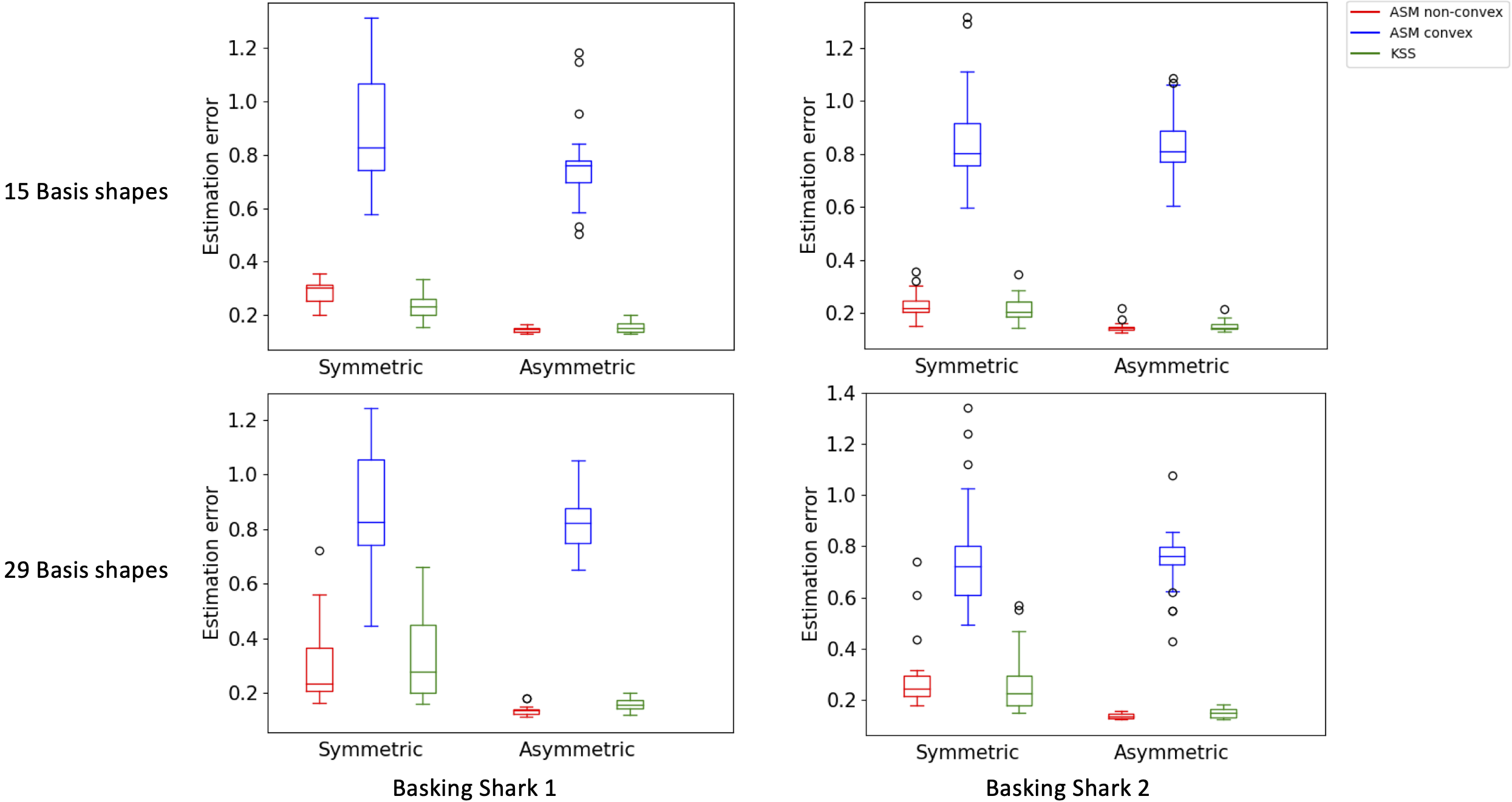}
    \caption{Boxplots of shape estimation errors of basking shark test shapes for the three approaches using the basis shapes from BS3.}
    \label{fig:bs_testing}
\end{figure}
\begin{table}[b]
    \parbox{\linewidth}{
        \centering
        \caption{Mean(variance) estimation errors for basking shark test shapes for symmetric and asymmetric projections using (15/29) basis shapes from Basking Shark 3.}
        \label{tab:bs_errors}
        \begin{tabular}{|c|c|c|c|c|c|}
        \hline
        &\#Basis Shapes & 15 (sym) & 15 (asym) &29 (sym) & 29 (asym)\\ \hline 
             \rule{0pt}{0.25cm} \multirow{3}{*}{BS1} & ASM NC & 0.231(0.002) & \textbf{0.146(3e-04)} & \textbf{0.306(0.020)} & \textbf{0.136(2e-04)}  \\
             & ASM Conv & 0.854(0.030) & 0.836(0.013) & 0.857(0.036) & 0.820(0.011) \\
             & KSS & \textbf{0.215(0.001)} & 0.151(3e-04) & 0.336(0.025) & 0.156(0.0003)\\ \hline
             \rule{0pt}{0.25cm} \multirow{3}{*}{BS2} & ASM NC & 0.286(0.001) & \textbf{0.145(1e-04)} & 0.281(0.014) & \textbf{0.137(9.7e-05)}\\
             & ASM Conv & 0.906(0.041) & 0.757 (0.021) & 0.755(0.041) & 0.747(0.013) \\
             & KSS & \textbf{0.236(0.001)} & 0.153(3e-04) & \textbf{0.256(0.012)} & 0.150(3.5e-04) \\
        \hline
        \end{tabular}
    }
\end{table}

Similar to human pose estimation, quantitative and qualitative analyses were performed for all approaches.
In the quantitative experiments, we used the head skeleton shapes (29 in total) of one of the basking sharks to estimate the shapes (28/29) of the other two.
Two types of projections, bilaterally symmetric and asymmetric, are used to estimate the 3D shape.
The former is the case when viewing the shark anteriorly (from the front) and the latter when viewing from the side (e.g.\ where one side of the throat is more visible than the other).
Both scenarios are equally important when extracting the 3D shape from images.
Fig.~\ref{fig:bs_testing} shows the errors in estimation using all the approaches and Table~\ref{tab:bs_errors} provides some error statistics. Again, statistically significant differences have been confirmed via Friedmann tests ($p < 0.001$) for each combination of subject and viewpoints.
Contrary to the human pose estimation, the ASM non-convex and our approach provide similar estimation errors for the basking shark experiment with the former being slightly superior for 5 out of the 8 cases.
These results indicate that the basking shark data might be located in a region of relatively low curvature in shape space that is well approximated by a hyper-plane.

\subsubsection{Qualitative validation on underwater imagery}

\begin{figure}[t]
    \centering
    \includegraphics[width=9cm]{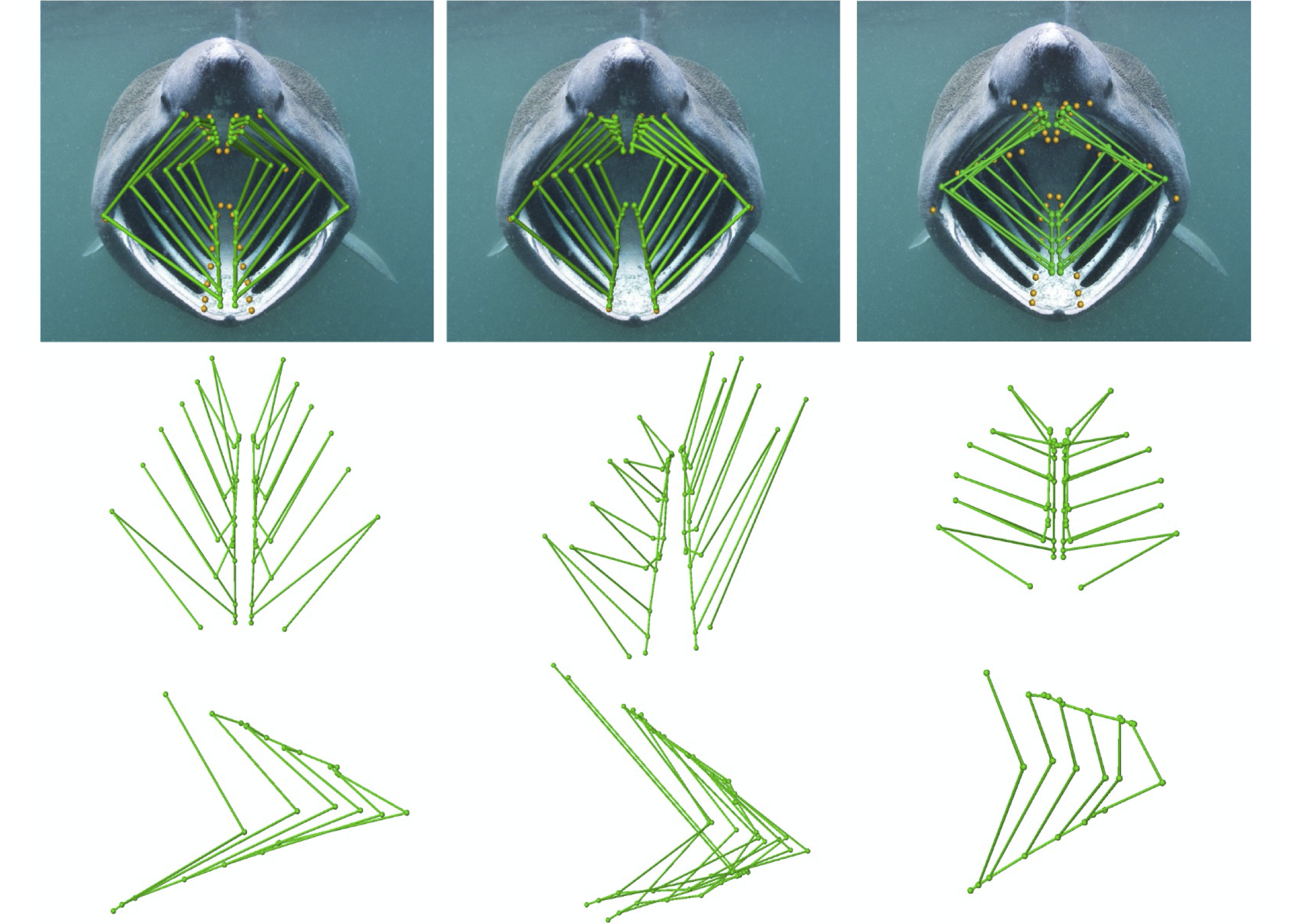}
    \caption{Comparison of the three approaches for 3D shape estimation applied to an under-water image\protect\footnotemark\ of a basking shark. Top row: Fitting of the estimated shapes (green) to the landmarks (yellow). Middle row: top views of the estimated shapes. Bottom row: side views. Left: ASM non-convex approach. Middle: ASM convex approach. Right: KSS approach (ours).}
    \label{fig:bs1}
\end{figure}
\footnotetext{Source: \url{https://www.amustard.com/library/albums/1/AMU_12_300611_0014.jpg}}
\begin{figure}[t]
    \centering
    \includegraphics[width=9cm]{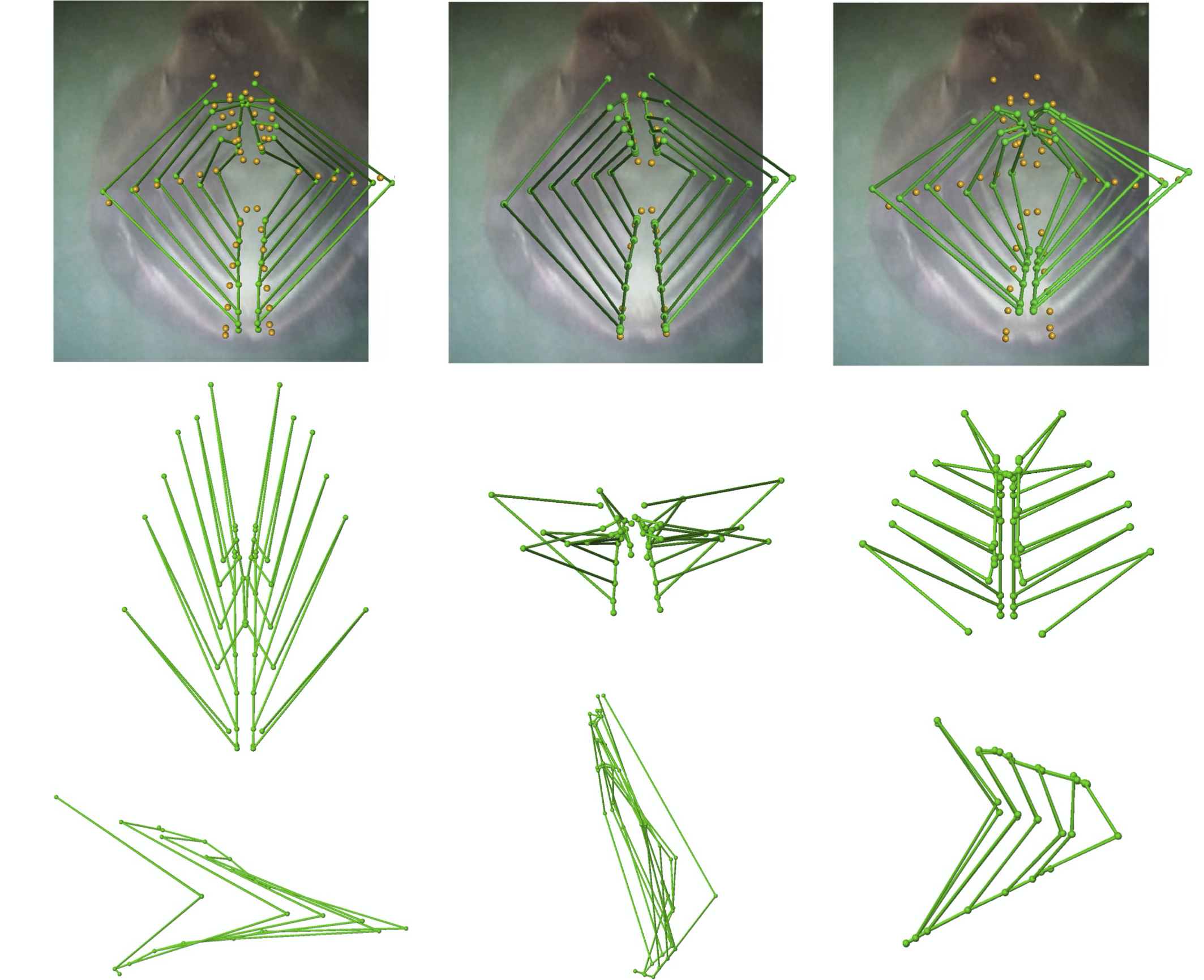}
    \caption{Comparison of the three approaches for 3D shape estimation applied to an under-water image\protect\footnotemark\ of a basking shark. Top row: Fitting of the estimated shapes (green) to the landmarks (yellow). Middle row: top views of the estimated shapes. Bottom row: side views. Left: ASM non-convex approach. Middle: ASM convex approach. Right: KSS approach (ours).}
    \label{fig:bs2}
\end{figure}
\footnotetext{Source: \url{https://www.manxbaskingsharkwatch.org/shark-files/biology-behaviour/feeding}}

We investigated the performances of the approaches in estimating the 3D structure of basking shark heads in underwater images with manually placed landmarks.
Shapes from all three sharks (86 in total) were used as basis shapes.
Figs.~\ref{fig:bs1} and \ref{fig:bs2} show the results obtained using the two ASM approaches and the novel KSS approach.
Similarly to the results for the estimation of real-world human poses, the shape estimations obtained from the two ASM approaches fit the 2D landmarks very well but fail to produce convincing results when looking from a different view.
This is particularly striking for the ASM convex approach but is also noticeable for the ASM non-convex method---the estimated shape of which is unnaturally elongated.
In contrast, applying the KSS approach results in 3D shapes that are very plausible and show a widely opened mouth.